\documentclass[letterpaper, 10 pt, conference]{ieeeconf}  

\IEEEoverridecommandlockouts                              

\usepackage{graphics} 
\usepackage{epsfig} 
\usepackage{amsmath} 
\usepackage{amssymb}  
\usepackage[caption=false]{subfig}
\usepackage{gensymb}
\usepackage{color}
\usepackage{url}
\usepackage{adjustbox}
\usepackage{array}
\usepackage{pifont}
\newcommand{\cmark}{\ding{51}}%
\usepackage{booktabs}
\usepackage[normalem]{ulem}
\usepackage{footnote}
\makesavenoteenv{tabular}
\makesavenoteenv{table}

\newcommand{\norm}[1]{\left\lVert #1 \right\rVert}

\title{\LARGE \bf
6D Object Pose Regression via Supervised Learning on Point Clouds
}

\author{Ge Gao$^{1}$, Mikko Lauri$^{1}$, Yulong Wang$^{2}$, Xiaolin Hu$^{2}$, Jianwei Zhang$^{1}$ and Simone Frintrop$^{1}$
\thanks{*This work is partially funded by the German Science Foundation (DFG) in project Crossmodal Learning, TRR 169,
and the National Natural Science Foundation of China (NSFC) under Grant No. 61621136008.}
\thanks{$^{1}$Department of Informatics, University of Hamburg, Germany
        {\tt\small \{gao,lauri,zhang\}@informatik.uni-hamburg.de}
        {\tt\small frintrop@informatik.uni-hamburg.de}}%
\thanks{$^{2}$State Key Laboratory of Intelligent Technology and Systems,
Department of Computer Science and Technology,
Tsinghua University, China 
        {\tt\small \{wang-yl15,xlhu\}@mails.tsinghua.edu.cn}}%
}

\begin{document}

\maketitle
\thispagestyle{empty}
\pagestyle{empty}

\begin{abstract}

This paper addresses the task of estimating the 6 degrees of freedom pose of a known 3D object from depth information represented by a point cloud.
Deep features learned by convolutional neural networks from color information have been the dominant features to be used for inferring object poses, while depth information receives much less attention. 
However, depth information contains rich geometric information of the object shape, which is important for inferring the object pose. 
We use depth information represented by point clouds as the input to both deep networks and geometry-based pose refinement and use separate networks for rotation and translation regression.
We argue that the axis-angle representation is a suitable rotation representation for deep learning, and use a geodesic loss function for rotation regression.
Ablation studies show that these design choices outperform alternatives such as the quaternion representation and L2 loss, or regressing translation and rotation with the same network.
Our simple yet effective approach clearly outperforms state-of-the-art methods on the YCB-video dataset.

\end{abstract}

\section{Introduction}
\label{sec:introduction}
  
The problem of 6 degrees of freedom (6D) object pose estimation is to determine the transformation from a local object coordinate system to a reference coordinate system (e.g.,\ camera or robot coordinate)~\cite{krulliccv15}.
The transformation is composed of 3D location and 3D orientation.
Robust and accurate 6D object pose estimation is of primary importance for many robotic applications such as grasping and dexterous manipulation.
The recent success of convolutional neural networks (CNNs) in visual recognition has inspired methods that use deep networks for learning features from RGB images~\cite{kehliccv17,oberwegereccv18,tremblaycorl18}.
These learned features are used for inferring 6D object poses.
Similarly, CNNs can also be applied to RGB-D images and treat depth information as an additional channel for feature learning~\cite{kehleccv16,zakharoviros17,lieccv18,buiicra18}.
However, in some scenarios, color information may not be available, and depth information is not in the 2-dimensional matrix format (e.g., laser range finder data), which can be easily processed with CNN-based systems.
Depth information can also be represented by a point cloud, which is an unordered set of points in a metric 3D space.
In existing methods, point clouds are mainly used for pose refinement~\cite{rusuicra09,sahin17,xiangrss18,jafariaccv18} or template matching with hand-crafted features extracted from point clouds~\cite{drostcvpr00,hinterstoissereccv16}.
Using point clouds in the registration stage confines its usage scope and hand-crafted features are usually less effective compared to deep-learned features.

In this work, we investigate how to accurately estimate the 6D pose of a known object represented by a point cloud segment containing only geometric information using deep networks.
Our approach is inspired by PointNet~\cite{qicvpr17}, a deep network for object classification and segmentation operating on point clouds. 
We adapt the system to the problem of pose estimation.
PointNet provides a method to apply deep learning to unordered point sets, and it is a suitable architecture for our purpose.
Our 6D pose regression method can be applied to any type of range data, e.g., data from laser range finders.

For developing our system, we investigate three open questions.
The first is how to efficiently use depth information in a deep learning-based system.
Although it has been shown by many applications that CNNs can extract powerful features from RGB-D information for specific tasks, due to the inherent difference between color and depth information, it is unclear whether this is an efficient way to treat depth information.
We argue that a point cloud is a more suitable structure and should be used in the scope of both deep networks and geometry-based optimization. 

The second question is whether translation and orientation should be estimated with separate networks or a single network in a supervised learning system.
During a supervised learning process, a network learns the mapping from its input to the desired output guided by a loss function.
Since the metric units for translation and orientation are different (i.e.,\ meters and radians), we argue that regressing them using separate networks and loss functions is a more suitable choice.
Our experiments show that an architecture with separate networks outperforms those with shared layers.

Another question is the choice of rotation representation and the loss function for measuring the distance between two rotations.
Quaternions have been a popular choice for many learning-based systems~\cite{xiangrss18,wangarxiv19}.
However, quaternions have the unit-norm constraint which imposes a limit on the network output range.
We argue that axis-angle is a more suitable choice because it is a constraint-free representation.
Concerning loss functions, L2 loss is a popular choice for measuring the distance between two rotations~\cite{wangarxiv19,kendall2017}.
We argue that the geodesic distance is a more suitable choice since it is well justified mathematically, and provides a clearer learning goal compared to the L2 loss.
We show experimentally that our arguments are valid.

\begin{figure*}[t!]
  \centering
    \includegraphics[width=0.7\textwidth]{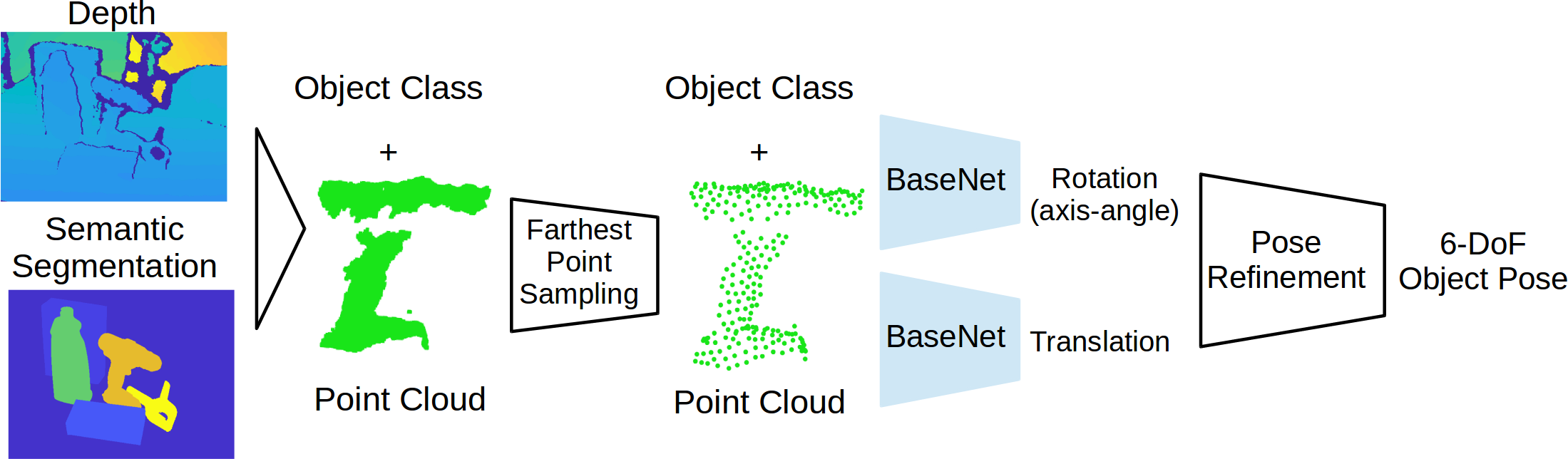}
    \setlength{\belowcaptionskip}{-10pt}
  \caption{System overview. A point cloud is created using the depth data and the output from a semantic segmentation method. This segment is processed with farthest point sampling to obtain a down-sampled segment with consistent surface structure.
  The segment with object class information is fed into two networks for rotation and translation prediction.
  The geometry-based iterative closest point algorithm is used for pose refinement.}
  \label{fig:system_fig}
\end{figure*}

Our contributions are thus as follows:
\begin{itemize}
  \item We present a simple yet effective system that infers the 6D pose of known objects represented by point cloud segments containing geometric information.
  This system is based on PointNet and exploits the point cloud structure by utilizing it as the input to both deep networks and an iterative closest point (ICP) refinement process.
  To the best of our knowledge, this is the first deep learning architecture based system that regresses 6D poses from only unordered point sets.
  \item We demonstrate that the proposed method outperforms state-of-the-art methods on a public dataset.
  Experimental results show that our system outperforms methods that use both color and depth information during pose inference stage. 
  This evaluation result indicates that the proposed system is an efficient way to use depth information for pose estimation.
  \item Ablation studies provide an evaluation of each system component.
  We show experimentally that each design choice has an impact on system performance. 
\end{itemize}


\section{Related Work} 
\label{sec:related_work}
Pose estimation has been well studied in the past using both color~\cite{kehliccv17,oberwegereccv18,tremblaycorl18} and depth data~\cite{hinterstoissericcv11,kouskouridastpami18}.
Since in this work we focus on investigating how to use geometric information during the pose inference stage, we mainly review works that use depth information.

 

With the common usage of depth cameras and laser range finders (e.g. LiDAR), methods using depth information have been proposed~\cite{hinterstoissericcv11,tejanieccv14,brachmanneccv14,krulliccv15,michelcvpr17}.
LINE-MOD~\cite{hinterstoissericcv11,hinterstoisseraccv12} is one of the first works that use hand-crafted features from depth information for pose estimation.
It uses surface normals as part of its local patch features.
This patch representation is adapted and used with random forest in~\cite{tejanieccv14,kouskouridastpami18}.
Surface normal is also used as an additional modality in~\cite{zakharoviros17}.
Another way of using depth information is to treat it as an extra image depth channel (RGBD) and feed it into a CNN~\cite{kehleccv16,balntasiccv2017,zakharoviros17,lieccv18,buiicra18}, or random forest~\cite{brachmanneccv14,krulliccv15,michelcvpr17} or a fully connected sparse autoencoder~\cite{doumanogloucvpr16} for feature extraction.
Depth is also used to create point clouds, which are used for generating pose hypotheses with 3D-3D correspondences and ICP refinement in~\cite{jafariaccv18}.
Point clouds are used to facilitate Point-to-Point matching in~\cite{drostcvpr00,hinterstoissereccv16}.
Geometry embeddings are extracted from point cloud segments with a deep network in~\cite{wangarxiv19}.
Approaches such as~\cite{xiangrss18} use color information to provide an initial pose estimate, then refine it with ICP using depth information.
Point cloud segments are also used for pose estimation in~\cite{qi2018cvpr}.
However, they only report experimental data for pose estimation for a single angle as opposed to all three as we do. 
Furthermore, they formulate rotation estimation as classification and discretize the rotation angles to bins.
It is not clear if this approach would scale up to three rotation angles.
In our previous work~\cite{gaoeccvw18}, we only considered rotation regression for single objects.
Here, we regress the full 6D pose in a multi-class setting.


For learning-based systems, \cite{xiangrss18} propose to predict translation and rotation with separate networks sequentially.
\cite{wangarxiv19} propose to regress translation and rotation with the same network.
Quaternions are used as the rotation representation for regression~\cite{xiangrss18,wangarxiv19,buiicra18}.
Bui et al. \cite{buiicra18} propose to use L2 loss function for rotation learning.
Axis-angle representation is also used in \cite{Mahendran2017cvprw}.
However, they only address estimating 3D rotation from RGB information, while we address both 3D rotation and 3D translation from point clouds.
Works that use deep networks for extracting features and do nearest neighbor search pose retrieval or classification into discretized bins to obtain object poses~\cite{kehleccv16,zakharoviros17,lieccv18,balntasiccv2017,qi2018cvpr} are not in the scope of this work and are not discussed in detail here.

Our work is most similar to DenseFusion~\cite{wangarxiv19} as we both use PointNet~\cite{qicvpr17} to extract features from point cloud segments.
However, there are two significant differences: first, during the pose estimation stage, we only use coordinate information from point clouds while DenseFusion also uses color information extracted by a CNN; second, we design regression targets and loss functions for rotation and translation individually while DenseFusion uses one regression target for the 6D pose.


\section{System Architecture} 
\label{sec:system_architecture}
Figure \ref{fig:system_fig} shows an overview of our system.
The proposed system for object 6D pose estimation is a multi-class system, i.e.,\ we use the same system to predict poses for objects from different classes.
Hence, an object segment, as well as the corresponding class information, is required as input to the pose estimation networks.
As semantic segmentation is a well-studied problem, we assume the object segment and class information is provided by an off-the-shelf method\footnote{Here, we use the semantic segmentation from ~\cite{xiangrss18}.} and focus on the object pose estimation from a point cloud segment in this work.
A point cloud segment is created using depth and the target object segment.
This segment is processed with Farthest Point Sampling (FPS)~\cite{eldarFPS97} to obtain a down-sampled segment with a consistent surface structure representation.
This down-sampled segment and class information are combined as the input for two separate networks for rotation estimation in axis-angle representation and translation prediction through translation residual regression.
The 6D pose is refined with a geometry-based optimization process to produce the final pose estimate.  

\begin{figure}[t]
\centering
\subfloat[]{\label{fig:basenet}\includegraphics[width = 0.75\columnwidth]{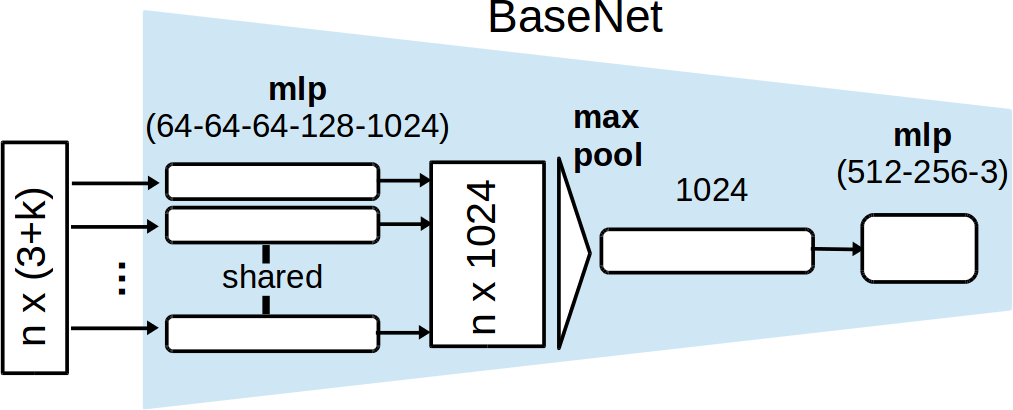}} \hspace{5mm}
\subfloat[]{\label{fig:network}\includegraphics[width = \columnwidth]{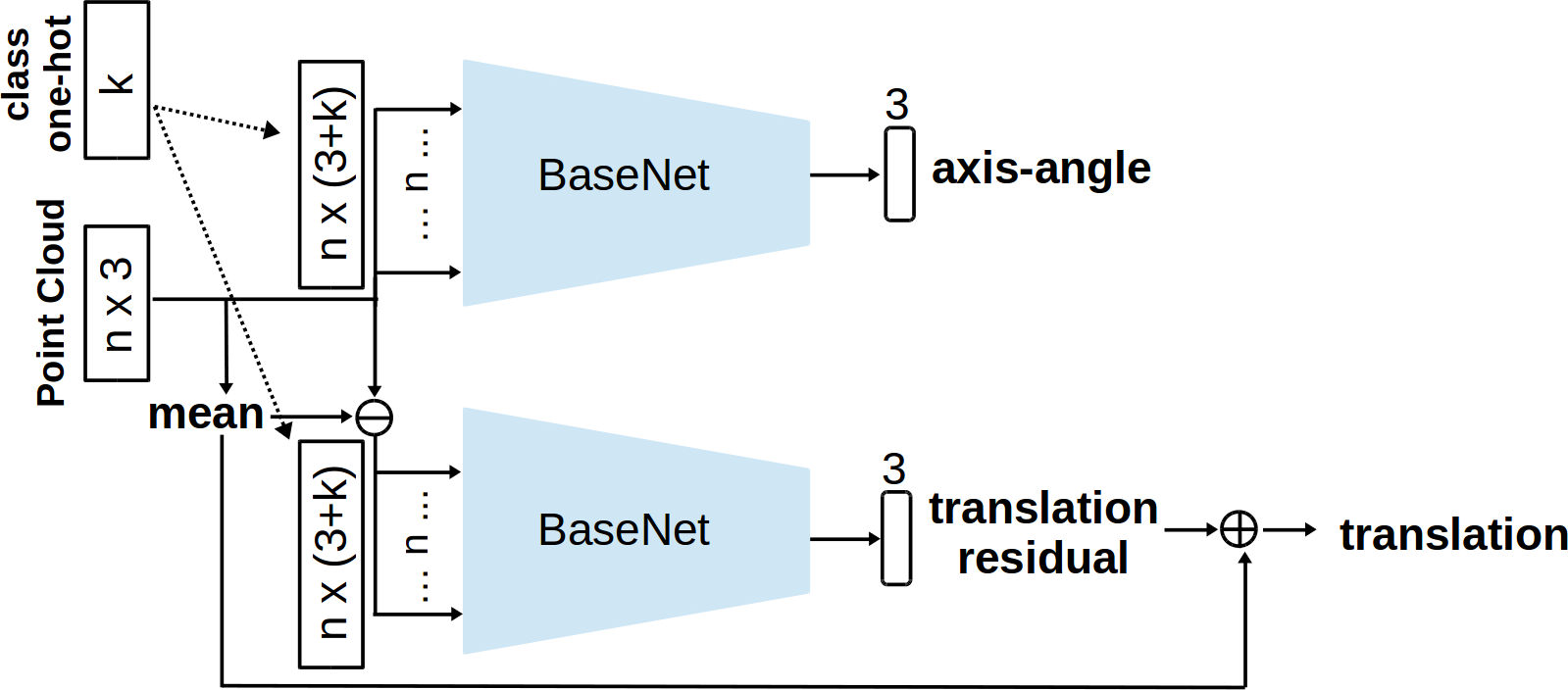}}
\setlength{\belowcaptionskip}{-10pt}
\caption{(a) The architecture of BaseNet. The numbers in parentheses are numbers of MLP layers. Numbers not in parentheses indicate the dimensions of intermediate feature vectors. A feature vector for each point is learned with shared weights. A max pooling layer aggregates the individual features into a global feature vector. A regression network with 3 fully-connected layers outputs the pose prediction. (b) Diagram for input and output of our pose networks. For the rotation network, the input is point coordinate information concatenated with class information per point, the output is rotation in axis-angle representation. For the translation network, the input coordinates are normalized by removing the mean. It outputs translation residual. The full translation is obtained by adding back the coordinate mean. The number of input points is $n$ and $k$ is the total number of classes.}
\label{fig:network_fig}
\end{figure}

Figure~\ref{fig:basenet} illustrates BaseNet, which is the basic building block of our system.
BaseNet is an adapted version of PointNet~\cite{qicvpr17}.
Given a point cloud with $n$ points as input, PointNet is invariant to all $n!$ possible permutations.
Each point is processed independently using multi-layer perceptrons (MLPs) with shared weights.
Compared to PointNet, we remove the spatial transformer blocks and adapt the dimension for the output layer to be $3$.
For each input point with class information, a feature vector is learned with shared weights.
These feature vectors are max-pooled to create a global representation of the input point cloud.
Finally, we use a three-layer regression MLP on top of this global feature to predict the poses.

Figure~\ref{fig:network} shows a more detailed diagram of our system.
We use two separate networks to handle translation and rotation estimation.
Input to the rotation network is a point cloud with $n$ points concatenated with the one-hot encoded class information.
In total, the input is a $n$ by $(3+k)$ array where $k$ is the total number of classes.
The output of the rotation network is the estimated rotation in axis-angle representation.
The translation network takes normalized point coordinates concatenated with class information and estimates the translation residual.
The full translation is obtained by adding back the coordinate mean.

\section{Supervised Learning for 6D Pose Regression}
\label{sec:supervised_learning_for_6d_pose_regression}
This section describes how the loss functions for 6D pose regression are formulated in our supervised learning framework.
Given a set of points $\mathcal{X} = \left\lbrace \mathbf{x}_i \in \mathbb{R}^3 \mid i = 1, \ldots, n \right\rbrace$ on the surface of a known object in the camera coordinate, the aim of pose estimation is to find a transformation that transforms $x_i$ from the object coordinates to the camera coordinates.
This transformation is composed of a translation and a rotation.
A translation consists of the displacements along the three coordinate axes. 
A rotation specifies the rotation around the three coordinate axes, and it has different representations such as axis-angle, Euler angles, quaternion, and rotation matrix. 
For supervised learning, suitable loss functions are required to measure the differences between predicted poses and ground truth poses.
For rotation learning, we argue that the axis-angle representation is the best suited for the learning task.
Geodesic distance is used as the loss function for rotation regression.
For translation learning, we predict the residual of translation.

\subsection{Rotation estimation with axis-angle and geodesic distance loss} 
\label{sub:rotation_estimation_in_axis_angle}
In the axis-angle representation, a vector $\mathbf{r}\in\mathbb{R}^3$ represents a rotation of $\theta = \norm{\mathbf{r}}_2$ radians around the unit vector $\frac{\mathbf{r}}{\norm{\mathbf{r}}_2}$~\cite{hartleyijcv13}.
Given an axis-angle representation $\mathbf{r} = \begin{bmatrix}r_1 & r_2 & r_3 \end{bmatrix}^T$, the corresponding rotation matrix $R$ is obtained via the exponential map $\exp:$
\begin{equation}
  R = \exp(\mathbf{r}_{\times}) = \mathrm{I}_{3\times3} + \frac{\sin\theta}{\theta}\mathbf{r}_{\times} + \frac{1-\cos\theta}{\theta^2} \mathbf{r}^2_{\times},
\end{equation}
where $\mathrm{I}_{3\times3}$ is the identity matrix and $\mathbf{r}_{\times}$ is the skew-symmetric matrix
\begin{equation}
\label{eq:skewsym}
  \mathbf{r}_{\times} =
 \begin{bmatrix}
  0 & -r_3 & r_2 \\
  r_3 & 0 & -r_1 \\
  -r_2 & r_1 & 0
 \end{bmatrix}.
\end{equation}

For rotation learning, we regress to a predicted rotation $\hat{\mathbf{r}}$.
Prediction $\hat{\mathbf{r}}$ is compared with ground truth rotation $\mathbf{r}$ via a \textit{rotation loss function} $l_r$, which is the geodesic distance between $\hat{R}$ and $R$~\cite{huynhjmiv09,hartleyijcv13}:
\begin{equation}
  l_r(\hat{\mathbf{r}}, \mathbf{r}) = \arccos\left(\frac{\mathrm{trace}(\hat{R}R^T)-1}{2}\right),
\end{equation}
where $\hat{R}$ and $R$ are the two rotation matrices corresponding to $\hat{\mathbf{r}}$ and $\mathbf{r}$, respectively.

This loss function directly measures the magnitude of rotation difference between $\hat{R}$ and $R$, so it is convenient to interpret.
Furthermore, the network can make constraint-free predictions with axis-angle representations, in contrast to e.g.\ quaternion representations which require normalization.

\subsection{Translation residual estimation}
\label{sub:translation_residual_estimation}
To simplify the learning task by reducing the variance in regression space for translation prediction, the learning target is chosen to be the residual of translation.
Given a translation residual $\widehat{\Delta\mathbf{t}}$, full translation prediction $\hat{\mathbf{t}}$ is obtained via
\begin{equation}
    \hat{\mathbf{t}} = \widehat{\Delta\mathbf{t}} + \mathbf{{\mu}_t},
\end{equation}
where $\mathbf{{\mu}_t}$ is the mean of $\mathcal{X}$.
$L2$-norm is used to measure the distance between prediction $\hat{\mathbf{t}}$ and ground truth $\mathbf{t}$, resulting in the \textit{translation loss function} $l_t(\hat{\mathbf{t}}, \mathbf{t})$:
\begin{equation}
    l_t(\hat{\mathbf{t}}, \mathbf{t}) = \norm{\mathbf{t} - \hat{\mathbf{t}}}_2.
\end{equation}

\subsection{Total loss function for 6D pose regression}

\noindent
The total loss is defined as the combination of the translation and the rotation loss:
\begin{equation}
    l(\mathbf{t},\hat{\mathbf{t}},\mathbf{r},\hat{\mathbf{r}}) =  \alpha l_t(\hat{\mathbf{t}}, \mathbf{t}) + l_r(\hat{\mathbf{r}}, \mathbf{r}),
\end{equation} 
where $\alpha$ is a scaling factor.
The total loss is used for training the pose estimation networks.


\section{Experiments} 
\label{sec:experiments}
We evaluate the proposed system on the YCB-Video dataset \cite{xiangrss18} and compare the performance with the state-of-the-art methods PoseCNN \cite{xiangrss18} and DenseFusion \cite{wangarxiv19}.
We also compare the performance on a subset of the object classes with a state-of-the-art RGB-based method DOPE~\cite{tremblaycorl18}.
Besides prediction accuracy and performance under occlusions, we also investigate the impact of using different network structures, as well as the influence of different rotation representations.
The implementation of our system is available on-line\footnote{\url{https://github.com/GeeeG/CloudPose}}.

\subsection{Experiment setup} 
\label{sub:experiment_setup}
The YCB video dataset~\cite{xiangrss18} contains 92 video sequences with total 133,827 frames of 21 objects selected from the YCB object set~\cite{calliram15} with 6D pose annotations.
We follow the official train/test split and use 80 video sequences for training.
Testing is performed on the $2,949$ key frames chosen from the remaining 12 sequences \cite{xiangrss18}.
80,000 frames of synthetic data are also provided by YCB-Video dataset as an extension to the training set.
During training, Adam optimizer is used with a learning rate of $0.0008$.
The batch size is $128$.
For the total loss, we use $\alpha=10$, which is given by the ratio between the expected error of translation and rotation at the end of the training \cite{kendalliccv15}.
The number of points of the input point cloud segment is $n=256$.
Batch normalization is applied to all layers.
No dropout is used.
All of our networks are trained for $90$ epochs.
For refinement, we use the Point-to-Point ICP registration provided by Open3D \cite{zhouarxiv18} and refine for 10 iterations.
The initial search radius is $0.01$m and is reduced by $10\%$ after each iteration.
For a fair comparison, all methods use object segmentation provided by PoseCNN during testing.

\subsection{Evaluation metrics} 
\label{sub:evaluation_metrics}
We use the average distance (AD) of model points  and the average distance for a rotationally symmetric object (ADS) proposed in~\cite{hinterstoisseraccv12} as evaluation metrics.
Given a 3D model represented as a set $\mathcal{M}$ with $m$ points, ground truth rotation $R$ and translation $\mathbf{t}$, as well as estimated rotation $\hat{R}$ and translation $\mathbf{\hat{t}}$, the AD is defined as:
\begin{align}
\mathrm{AD}=\frac{1}{m}\displaystyle\sum_{\mathbf{x}\in\mathcal{M}} \norm{ (R\mathbf{x}+\mathbf{t})-(\hat{R}\mathbf{x}+\hat{\mathbf{t}}) }_2.
\end{align}
ADS is computed using closest point distance.
It provides a distance measure that considers possible pose ambiguities caused by rotational symmetry:
\begin{align}
\mathrm{ADS}=\frac{1}{m}\displaystyle\sum_{\mathbf{x}_1\in\mathcal{M}} \min_{\mathbf{x}_2\in\mathcal{M}}\norm{ (R\mathbf{x}_1+\mathbf{t})-(\hat{R}\mathbf{x}_2+\hat{\mathbf{t}}) }_2.
\end{align}

A 6D pose estimate is considered to be correct if AD and ADS are smaller than a given threshold.
We report the area under error threshold-accuracy curve (AUC) for AD and ADS.
The maximum thresholds for both curves are set to $0.1$m.
Furthermore, we also provide ADS accuracy with threshold $0.01$m ($<$1cm) to illustrate the performance accuracy under a smaller error tolerance.

\subsection{Prediction accuracy} 
\label{sub:comparison_the_state_of_the_arts}
Evaluation results averaged for all 21 objects in the YCB-Video dataset are shown in Table \ref{tab:add}.
PoseCNN~\cite{xiangrss18} uses RGB information to provide an initial pose estimate (PC w/o ICP), then uses depth information with a highly customized ICP for pose refinement (PC).
DenseFusion~\cite{wangarxiv19} (DF) uses both color and point cloud features extracted by deep networks to give per-pixel pose estimate for final pose voting, and iterative pose refinement is performed with an extra network module.
Ours w/o ICP is the estimated pose from the proposed system architecture (Section \ref{sec:system_architecture}), and Ours is the result after ICP refinement.
We also perform the ICP refinement on DF results (DF+ICP).
For the overall performance in Table \ref{tab:add}, we highlight the best performance in bold font.
Details regarding the data type used by pose regression networks and the post process are also presented.

Our method achieves state-of-the-art performance using only depth information.
In terms of AD, we outperform both PC an DF.
We observe that DF+ICP shows small improvement compared to DF.
One possible reason is the sensitivity of ICP to the initial pose guess, if the method already performs well without refinement, ICP is able to provide further gains.
If the initial guess is poor, ICP can even make the results worse.
This result indicates that features learned from depth information represented by unordered point clouds are sufficient for accurately regressing 6D pose.
Furthermore, this also shows that the proposed approach is an efficient way to use depth information in a deep learning framework for pose regression.

Performance for individual objects is shown in Table \ref{tab:add_details}.
We use the trained network for six objects provided by the authors of DOPE~\cite{tremblaycorl18} and report the results.
The AD results are not available because the object coordinate frames used in the YCB object dataset~\cite{calliram15}, YCB video dataset for PoseCNN~\cite{xiangrss18} and DOPE are different.
As our method uses the frames from \cite{xiangrss18}, and the transformation between \cite{calliram15} and \cite{xiangrss18} is not publicly available, we can not find the correspondence between model points required for AD.
We also applied ICP to DOPE pose estimates, but the performance was not improved.
A possible reason is the sensitivity of ICP to the initial pose estimate.

\begin{figure*}[hbt!]
\centering
\begin{tabular}{cccccc}
PC~\cite{xiangrss18} & DF~\cite{wangarxiv19} & Ours & PC~\cite{xiangrss18} & DF~\cite{wangarxiv19} & Ours \\

\subfloat{\includegraphics[width = .144\textwidth]{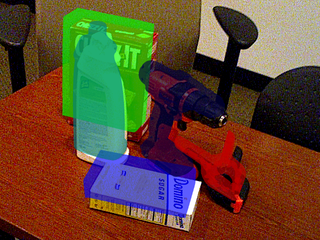}}&
\subfloat{\includegraphics[width = .144\textwidth]{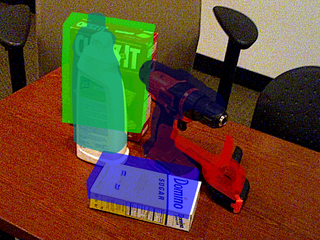}}&
\subfloat{\includegraphics[width = .144\textwidth]{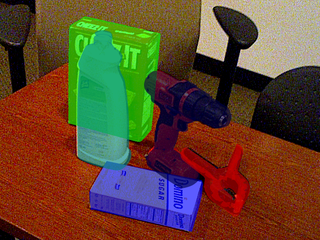}}& 
\subfloat{\includegraphics[width = .144\textwidth]{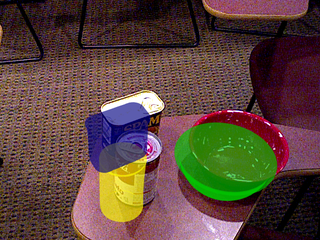}}&
\subfloat{\includegraphics[width = .144\textwidth]{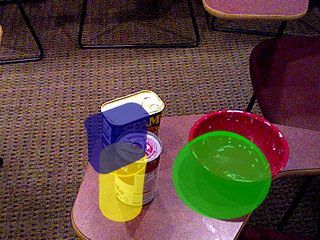}}&
\subfloat{\includegraphics[width = .144\textwidth]{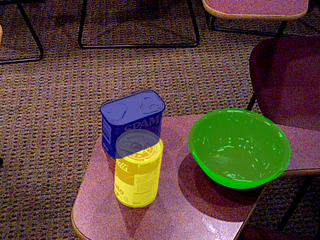}}\\ 

\subfloat{\includegraphics[width = .144\textwidth]{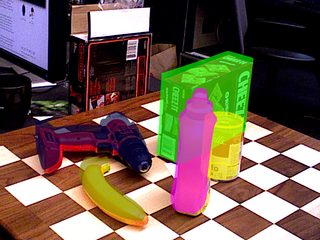}}&
\subfloat{\includegraphics[width = .144\textwidth]{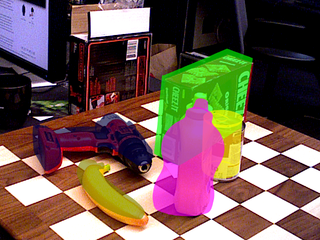}}&
\subfloat{\includegraphics[width = .144\textwidth]{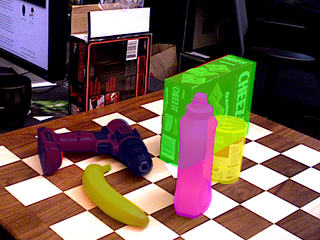}}&
\subfloat{\includegraphics[width = .144\textwidth]{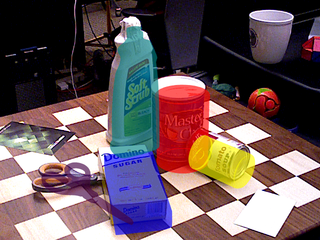}}&
\subfloat{\includegraphics[width = .144\textwidth]{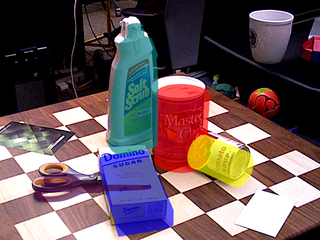}}&
\subfloat{\includegraphics[width = .144\textwidth]{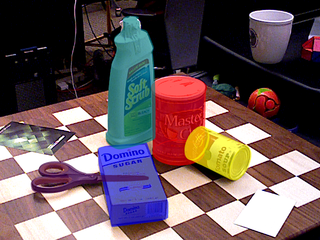}}
 
\end{tabular}
\caption{Qualitative results for 6D pose estimation. From left to right: PoseCNN (PC)~\cite{xiangrss18}, DenseFusion (DF)~\cite{wangarxiv19}, and ours. The colored overlay indicates the predicted pose of the target object.
Our method gives more accurate translation estimates, and also is able to give accurate rotation estimation for texture-less object (e.g. the red bowl).
More qualitative results are available in the supplementary video.}
\label{fig:result_visual}
\end{figure*}

Some qualitative results are shown in Figure \ref{fig:result_visual}.
Pose estimates from PC, DF and our method are used for projecting object models onto 2D images.
More qualitative results are available in the supplementary video.


\begin{table}[]
\caption{Quantitative evaluation of 6D pose on the YCB-Video Dataset~\cite{xiangrss18}. Best performance is in bold font.}
\label{tab:add}
\centering
\resizebox{\columnwidth}{!}{
\begin{tabular}{@{}lcccccc@{}}
\toprule
            & RGB & Depth & ICP & AD            & ADS           & \textless{}1 cm \\ \midrule
PC w/o ICP~\cite{xiangrss18}   & \cmark   &       &     & 51.5          & 75.6          & 26.1            \\
PC~\cite{xiangrss18}          & \cmark   & \cmark     & \cmark   & 77.8          & 93.6          & 88.4            \\
DF~\cite{wangarxiv19}          & \cmark   & \cmark     &     & 74.7          & 93.9          & 87.6            \\
DF~\cite{wangarxiv19} + ICP\footnote{we apply ICP to DF results after iterative refinement.}   & \cmark   & \cmark     & \cmark   & 76.3          & \textbf{94.7} & 89.0              \\
Ours w/o ICP &     & \cmark    &     & 76.0            & 91.3          & 80.9            \\
Ours        &     & \cmark     & \cmark   & \textbf{82.7} & \textbf{94.7} & \textbf{90.3}   \\ \bottomrule
\end{tabular}
}
\end{table}

\begin{table}[t]
\caption{Pose estimation accuracy per object class on the YCB-Video Dataset~\cite{xiangrss18}. Best per class performance for AD(S) is in bold font.
Ours achieves the best performance on a majority of object classes.}
\label{tab:add_details}
\setlength\tabcolsep{3pt}
\resizebox{\columnwidth}{!}{%
\begin{tabular}{@{}lp{0.4cm}cccccccccc@{}}
\toprule
                                           & \multicolumn{2}{c}{DOPE~\cite{tremblaycorl18}}         & \multicolumn{3}{c}{PC~\cite{xiangrss18}}                                    & \multicolumn{3}{c}{DF~\cite{wangarxiv19}}                                    & \multicolumn{3}{c}{Ours}             \\ \midrule
\multicolumn{1}{l|}{}                      & ADS  & \multicolumn{1}{c|}{\textless{}1cm} & AD            & ADS           & \multicolumn{1}{c|}{\textless{}1cm} & AD            & ADS           & \multicolumn{1}{c|}{\textless{}1cm} & AD            & ADS           & \textless{}1cm \\ \midrule
\multicolumn{1}{l|}{02\_master\_chef}   & ---  & \multicolumn{1}{c|}{---}  & 68.1          & 95.8          & \multicolumn{1}{c|}{99.5} & \textbf{73.2} & \textbf{96.4} & \multicolumn{1}{c|}{100}  & 46.9          & 95.4          & 95.4 \\
\multicolumn{1}{l|}{03\_cracker\_box}       & 62.7 & \multicolumn{1}{c|}{29.6} & 83.4          & 92.7          & \multicolumn{1}{c|}{84.8} & \textbf{94.2} & \textbf{95.8} & \multicolumn{1}{c|}{97}   & 76.7          & 93            & 80.4 \\
\multicolumn{1}{l|}{04\_sugar\_box}         & 85.0 & \multicolumn{1}{c|}{33.4} & 97.1          & 98.2          & \multicolumn{1}{c|}{100}  & 96.5          & 97.6          & \multicolumn{1}{c|}{100}  & \textbf{97.5} & \textbf{98.5} & 99.7 \\
\multicolumn{1}{l|}{05\_tomato\_soup}   & 88.5 & \multicolumn{1}{c|}{74.5} & 83.6          & \textbf{96.6} & \multicolumn{1}{c|}{99}   & \textbf{87.4} & \textbf{96.6} & \multicolumn{1}{c|}{99.1} & 72.7          & 96.5          & 96.8 \\
\multicolumn{1}{l|}{06\_mustard\_bottle}    & 90.7 & \multicolumn{1}{c|}{65.3} & \textbf{98}   & \textbf{98.6} & \multicolumn{1}{c|}{98.9} & 94.8          & 97.3          & \multicolumn{1}{c|}{97.8} & 79.2          & 97.7          & 94.1 \\
\multicolumn{1}{l|}{07\_tuna\_fish\_can}     & ---  & \multicolumn{1}{c|}{---}  & \textbf{83.9} & 97.1          & \multicolumn{1}{c|}{97.6} & 81.8          & 97.1          & \multicolumn{1}{c|}{99.5} & 72            & \textbf{97.7} & 100  \\
\multicolumn{1}{l|}{08\_pudding\_box}       & ---  & \multicolumn{1}{c|}{---}  & \textbf{96.6} & \textbf{97.9} & \multicolumn{1}{c|}{100}  & 93.2          & 95.9          & \multicolumn{1}{c|}{98.6} & 94.4          & 97.3          & 91.1 \\
\multicolumn{1}{l|}{09\_gelatin\_box}       & 84.6 & \multicolumn{1}{c|}{36.9} & 98.1          & 98.8          & \multicolumn{1}{c|}{100}  & 96.7          & 98            & \multicolumn{1}{c|}{100}  & \textbf{98.6} & \textbf{99}   & 100  \\
\multicolumn{1}{l|}{10\_potted\_meat}   & 32.0 & \multicolumn{1}{c|}{3.7}  & 86            & 94.3          & \multicolumn{1}{c|}{87.5} & 87.8          & 95            & \multicolumn{1}{c|}{92}   & \textbf{90.6} & \textbf{95.7} & 93.7 \\
\multicolumn{1}{l|}{11\_banana}            & ---  & \multicolumn{1}{c|}{---}  & 91.9          & 97.1          & \multicolumn{1}{c|}{95}   & 83.6          & 96.2          & \multicolumn{1}{c|}{98.2} & \textbf{95.1} & \textbf{97.7} & 95.5 \\
\multicolumn{1}{l|}{19\_pitcher\_base}      & ---  & \multicolumn{1}{c|}{---}  & \textbf{96.9} & 97.8          & \multicolumn{1}{c|}{99.6} & 96.6          & 97.5          & \multicolumn{1}{c|}{99.5} & 96.1          & \textbf{97.9} & 100  \\
\multicolumn{1}{l|}{21\_bleach\_clean}   & ---  & \multicolumn{1}{c|}{---}  & 92.5          & 96.9          & \multicolumn{1}{c|}{95.1} & 89.7          & 95.8          & \multicolumn{1}{c|}{99.4} & \textbf{95.4} & \textbf{97.4} & 98.4 \\
\multicolumn{1}{l|}{24\_bowl}              & ---  & \multicolumn{1}{c|}{---}  & 14.4          & 81            & \multicolumn{1}{c|}{42.9} & 5.9           & 89.5          & \multicolumn{1}{c|}{55.7} & \textbf{83.9} & \textbf{97.7} & 99.3 \\
\multicolumn{1}{l|}{25\_mug}               & ---  & \multicolumn{1}{c|}{---}  & 81.1          & 94.9          & \multicolumn{1}{c|}{97.6} & 88.8          & 96.7          & \multicolumn{1}{c|}{98}   & \textbf{93.9} & \textbf{97.8} & 99.7 \\
\multicolumn{1}{l|}{35\_power\_drill}       & ---  & \multicolumn{1}{c|}{---}  & \textbf{97.7} & \textbf{98.2} & \multicolumn{1}{c|}{99.3} & 93            & 96.1          & \multicolumn{1}{c|}{97.8} & 94.9          & 97.7          & 96.7 \\
\multicolumn{1}{l|}{36\_wood\_block}        & ---  & \multicolumn{1}{c|}{---}  & 70.9          & 87.6          & \multicolumn{1}{c|}{74.4} & 30.9          & 92.8          & \multicolumn{1}{c|}{88.8} & \textbf{90}   & \textbf{94.9} & 97.5 \\
\multicolumn{1}{l|}{37\_scissors}          & ---  & \multicolumn{1}{c|}{---}  & \textbf{78.4} & \textbf{91.7} & \multicolumn{1}{c|}{68}   & 77.4          & 91.9          & \multicolumn{1}{c|}{71.3} & 75.8          & 91.3          & 63   \\
\multicolumn{1}{l|}{40\_large\_marker}      & ---  & \multicolumn{1}{c|}{---}  & 85.3          & 97.2          & \multicolumn{1}{c|}{97.1} & \textbf{93}   & 97.6          & \multicolumn{1}{c|}{100}  & 92.2          & \textbf{98}   & 100  \\
\multicolumn{1}{l|}{51\_large\_clamp}       & ---  & \multicolumn{1}{c|}{---}  & 52.2          & 75.3          & \multicolumn{1}{c|}{67.4} & 26.4          & 72.6          & \multicolumn{1}{c|}{33.3} & \textbf{68.5} & \textbf{77.4} & 69.6 \\
\multicolumn{1}{l|}{52\_e\_large\_clamp} & ---  & \multicolumn{1}{c|}{---}  & 25.9          & 74.9          & \multicolumn{1}{c|}{48.2} & 16.6          & 77.4          & \multicolumn{1}{c|}{10.9} & \textbf{25.3} & 66.4          & 22   \\
\multicolumn{1}{l|}{61\_foam\_brick}        & ---  & \multicolumn{1}{c|}{---}  & 48.1          & 97.2          & \multicolumn{1}{c|}{99.7} & 59            & 92            & \multicolumn{1}{c|}{100}  & \textbf{92.9} & \textbf{98}   & 99.3 \\ \bottomrule
\end{tabular}
}
\end{table}



\subsection{Occlusion} 
\label{sub:occlusion}
\begin{figure}[t]
  \centering
    \includegraphics[width=\columnwidth]{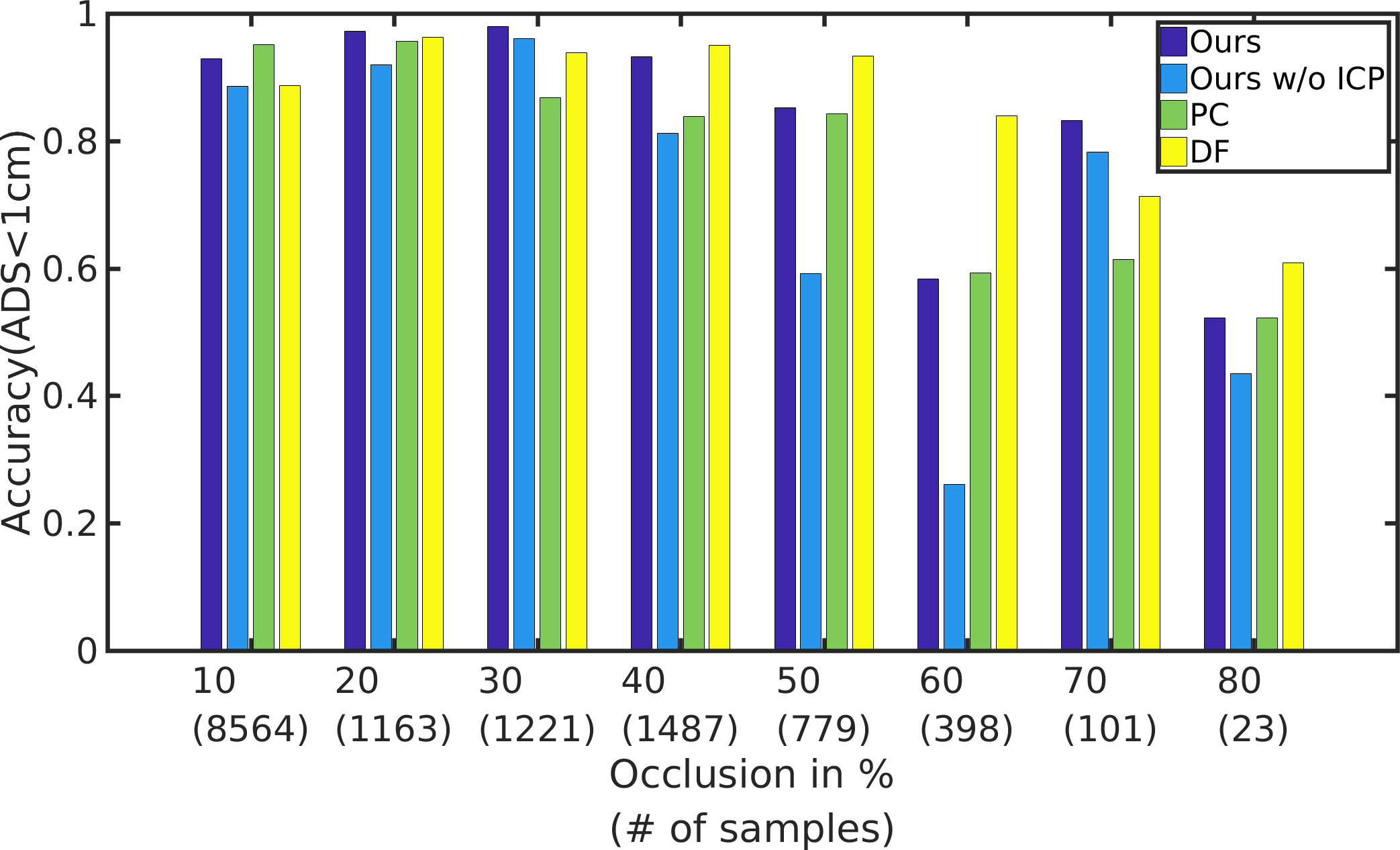}
    \caption{Effect of occlusion compared to PC~\cite{xiangrss18} and DF~\cite{wangarxiv19}. The horizontal axis denotes the upper limit (occlusion in $\%$) of each bin. Width for each bin is $10\%$. Numbers in parentheses denote the number of samples in corresponding bin.
    Ours is competitive with state-of-the-art methods when occlusion is lower than $40\%$.}
  \label{fig:occlusion_plot}
\end{figure}

For a given target object in a frame, the occlusion factor $O$ of the object is defined as~\cite{gaoeccvw18}
\begin{align}
O = 1 - \frac{\lambda}{\nu},
\end{align}
where $\lambda$ is the number of pixels in the 2D ground truth segmentation, and $\nu$ is the number of pixels in the projection of the 3D object model onto the image plane using the camera intrinsic parameters and the ground truth 6D pose, when we assume the object would be fully visible.
The occlusion factor of the YCB-Video dataset ranges from $0.8\%$ to $81\%$.
We divide this range into $8$ bins with a bin width of $10\%$ and report the prediction accuracy (ADS) with a threshold of $1$ cm.
Figure \ref{fig:occlusion_plot} illustrates the results.
It can be observed that our method (Ours) has competitive performance when the occlusion is lower than $40\%$, then both ours and PC start to suffer as the amount of occlusion increases.
One possible reason is that DF outputs per-pixel prediction with confidence scores, while ours and PC provide only one pose prediction.
This per-pixel prediction may have helped to provide better performance when the amount of occlusion is higher than $40\%$.

\subsection{Network architecture: ablation study} 
\label{sub:network_architecture}
To investigate whether translation and rotation should be regressed with the same or separate networks, we compare the performance of different architectures.
We alter the network architecture by incrementally sharing the layers between translation and rotation networks.
Table \ref{tab:diff_struct} shows the result in terms of AD, ADS, and accuracies for translation and rotation under certain thresholds.
None denotes the proposed architecture which regresses translation and rotation with two separate networks.
The numbers in the first column denote the number of shared layers between translation and rotation BaseNet (Figure \ref{fig:network_fig}).
We compare performance without ICP refinement.
When sharing layers, the performance is worse than using two separate networks.

We also tested an architecture that shares all the layers while having the same amount of parameters as the proposed structure with a doubled layer width.
The performance is similar to the architecture with the single width, and this verifies that the performance deterioration is not caused by insufficient network capacity.
This result verifies that using separated networks for translation and rotation is a more suitable design choice.

\begin{table}[t]
\centering
\caption{Accuracy with different network structures.
Best performance is in bold font.
Using two separate networks performs best.}
\label{tab:diff_struct}
\resizebox{0.85\columnwidth}{!}{%
\begin{tabular}{@{}ccccc@{}}
\toprule
shared layers & AD            & ADS           & rot\_err\textless{}10\degree & tran\_err\textless{}1 cm \\ \midrule
none & \textbf{76.0} & \textbf{91.3} & \textbf{41.3}               & \textbf{73.0}           \\
1             & 75.2          & 91.1          & 39.5                        & 72.0                    \\
2             & 75.5          & 91.1          & 38.6                        & 69.8                    \\
3             & 75.1          & 91.1          & 41.2                        & 69.6                    \\
4             & 75.5          & 91.2          & 39.4                        & 70.3                    \\
5             & 75.2          & 91.1          & 39.1                        & 69.7                    \\
all           & 63.0          & 87.8          & 23.3                        & 69.3                    \\ \bottomrule
\end{tabular}
}
\end{table}

\begin{table}[t]
\centering
\caption{Prediction accuracy with different rotation representations and loss functions. Best performance is in bold font.}
\label{tab:rot_and_loss}
\subfloat[Different rotation representations. Geodesic distance is used as the loss function for both cases.]{
\label{tab:rot_rep}
\resizebox{0.7\columnwidth}{!}{%
\begin{tabular}{@{}cccc@{}}
\toprule
Rotation representation & \textless{}10\degree & \textless{}15\degree & \textless{}20\degree \\ \midrule
Axis-angle    & \textbf{41.3}        & \textbf{52.7}        & \textbf{59.6}        \\
Quaternion    & 37.1                 & 48.5                 & 56.7                 \\ \bottomrule
\end{tabular}
}
}
\vspace{5mm}
\subfloat[Different loss functions for rotation regression. Axis-angle representation is used for both cases.]{
\label{tab:loss_func}
\resizebox{0.55\columnwidth}{!}{%
\begin{tabular}{@{}cccc@{}}
\toprule
Loss function & \textless{}10\degree & \textless{}15\degree & \textless{}20\degree \\ \midrule
Geodesic   & \textbf{41.3}        & \textbf{52.7}        & \textbf{59.6}        \\
L2         & 40.8                 & 52.5                 & 58.4                 \\ \bottomrule
\end{tabular}
}
}
\end{table}

\subsection{Rotation representation and loss function: ablation study}
We investigate the impact of different rotation representations and loss functions.
For comparing quaternion to axis-angle, we adapted our rotation network to have $4$-dimensional output instead of $3$.
The output is normalized and then converted to the axis-angle representation.
We use the same loss function as described in Section \ref{sec:supervised_learning_for_6d_pose_regression}.
For comparing L2 loss with Geodesic distance, we keep the rotation representation in axis-angle format and apply different loss functions.
Table \ref{tab:rot_and_loss} shows the accuracy of rotation prediction with different thresholds.
With the same loss function, using axis-angle yields a better result than quaternion.
This indicates that axis-angle is a better choice for rotation learning.
With the same rotation representation, L2 loss slightly underperforms geodesic loss.
Since geodesic distance also has a better mathematical justification, this makes it a better choice.


\subsection{Time performance} 
\label{sub:time_performance}
We measure the time performance on a Nvidia Titan X GPU.
The system is implemented with Tensorflow.
Pose estimation by a forward pass through our network takes 0.11 seconds for a single object. 
The 10 iterations of ICP refinement require an additional 0.3 seconds.

\section{Conclusion}
\label{sec:conclusion}
We propose a system for fast and accurate 6D pose estimation of known objects.
We formulate the problem as a supervised learning problem and use two separate networks for rotation and translation regression, and use point clouds as input for the regression.
We use axis-angle as rotation representation and geodesic distance as the loss function for rotation regression.
Ablation studies show that these design choices outperform the commonly used quaternion representation and L2 loss.
Experimental results show that the proposed system outperforms two state-of-the-art methods on a public benchmark.

To the best of our knowledge, this is the first deep learning system that regresses 6D object poses from only depth information represented by unordered point clouds.
Features extracted from point clouds with deep networks can be used for accurately regressing object pose.
Our pose regression system can be applied to range data from other sensors such as laser range finders.
In the future work, we will investigate aspects such as pose estimation for rotational symmetry objects using only geometric information.







\bibliographystyle{IEEEtran}
\bibliography{root}

\end{document}